# End-to-End Trainable One-Stage Parking Slot Detection Integrating Global and Local Information

Jae Kyu Suhr and Ho Gi Jung

*Abstract*— This paper proposes an end-to-end trainable one-stage parking slot detection method for around view monitor (AVM) images. The proposed method simultaneously acquires global information (entrance, type, and occupancy of parking slot) and local information (location and orientation of junction) by using a convolutional neural network (CNN), and integrates them to detect parking slots with their properties. This method divides an AVM image into a grid and performs a CNN-based feature extraction. For each cell of the grid, the global and local information of the parking slot is obtained by applying convolution filters to the extracted feature map. Final detection results are produced by integrating the global and local information of the parking slot through non-maximum suppression (NMS). Since the proposed method obtains most of the information of the parking slot using a fully convolutional network without a region proposal stage, it is an end-to-end trainable one-stage detector. In experiments, this method was quantitatively evaluated using the public dataset and outperforms previous methods by showing both recall and precision of 99.77%, type classification accuracy of 100%, and occupancy classification accuracy of 99.31% while processing 60 frames per second.

*Index Terms*— Parking slot detection, deep learning, convolutional neural network, end-to-end, one-stage detector

## I. INTRODUCTION

Automatic parking systems have been consistently researched as a key element of autonomous driving [1]. Vacant parking space detection is undoubtedly the first step of an automatic parking system. This task has been conducted in four approaches: free space-based, slot marking-based, user interface-based, and infrastructure-based [2]. Among them, the first two approaches have been more widely researched compared to the others. The free space-based approach detects vacant parking spaces by recognizing adjacent parked vehicles. It works well when the parked vehicles are in favorable positions, but its performance depends on the existence and poses of the parked vehicles. This drawback can be mitigated by the slot marking-based approach. This approach detects vacant parking spaces by recognizing parking slot markings on the ground. It can accurately detect parking spaces regardless of the existence and poses of the parked vehicles, but its performance depends on the visual condition of the parking slot markings. Most of the methods in this approach find lines, corners, or pixels of a specific color and combine them according to predetermined geometric constraints of parking slot markings [2]-[18].

Recently, deep learning-based object detection has been widely researched because of its impressive performance, and attempts to implement it using edge devices have been actively conducted [19]. Accordingly, it has also been applied to parking slot detection tasks and showed more robust and higher detection performance than traditional methods [20]-[22]. Deep learning-based parking slot detection methods can be categorized into two approaches. The first approach uses deep learning techniques along with traditional rule-based techniques [20], [21]. The methods in this approach first find junctions that make up entrances of parking slots using deep learning techniques and then pair them using manually designed geometric rules to generate parking slots. This approach can precisely estimate the locations of the parking slots based on the locations of the junctions detected by deep learning techniques, but it cannot be trained end-to-end due to the use of manually designed geometric rules. The second approach detects parking slots by applying the existing deep learning-based general object detector [22]. This method can be trained end-to-end, but it cannot estimate the precise location and orientation of the parking slot because the existing general object detector, which has a limitation in terms of the localization accuracy, is directly used without being specialized in parking slot detection tasks. In the viewpoint of the automatic parking system, the positioning accuracy of the detection result is significantly important because the vehicle should be controlled according to the detected position.

This paper proposes a novel method that can overcome the limitations of the previous deep learning-based parking slot detection methods. The proposed method simultaneously acquires global information (entrance, type, and occupancy of parking slot) and local information (location and orientation of junction) by using a convolutional neural network (CNN), and integrates them to find parking slots with their properties. Fig. 1

J. K. Suhr is with the School of Intelligent Mechatronics Engineering, Sejong University, Seoul 05006, South Korea (e-mail: jksuhr@sejong.ac.kr).
Ho Gi Jung is with the Department of Electronic Engineering, Korea National University of Transportation, Chungju 27469, South Korea (e-mail: hogijung@ut.ac.kr).

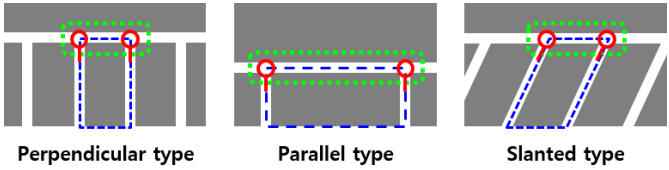

Fig. 1. Definitions of slot (blue dashed line), entrance (green dotted line), and junction (red solid line).

shows definitions of slot, entrance, and junction in cases of three types of parking slot markings. In this figure, blue dashed, green dotted, and red solid lines indicate slots, entrances, and junctions, respectively. The proposed method divides an around view monitor (AVM) image into a grid and performs CNN-based feature extraction. For each cell of the grid, global and local information of the parking slot is obtained by applying convolution filters to the extracted feature map. The global information consists of the entrance location, type, and occupancy of the parking slot that includes the center of each cell. The local information consists of the location and orientation of the junction that is included in each cell. Both the entrance location in the global information and the junction location in the local information represent the same location, but the junction location is more precise than the entrance location. This is because during the global information extraction, each cell is obliged to estimate the entrance located distant from it but during the local information extraction, each cell is obliged to estimate the junction located inside of it. Even though the entrance location is less precise than the junction location, it should be obtained because it contains pairing information of two junctions. Final detection results are produced by integrating the global and local information of the parking slots through the non-maximum suppression (NMS) based on junctions. Since the proposed method obtains most of the information of the parking slots using a fully convolutional network without a region proposal stage, it is an end-to-end trainable one-stage detector. In experiments, the proposed method was quantitatively evaluated using the public dataset and outperforms previous methods by showing both recall and precision of 99.77%, type classification accuracy of 100%, and occupancy classification accuracy of 99.31% while processing 60 frames per second. It also shows 1.02 pixels and 0.18° for location and orientation errors, respectively.

The proposed method has the following contributions over the previous deep learning-based parking slot detection methods:
1) It can achieve both high detection rate and positioning accuracy by integrating the global and local information of the parking slot.
2) It can be trained end-to-end and rapidly detect parking slots because it uses a fully convolutional network and one-stage detection strategy.
3) It does not require inconvenient procedures for setting geometric rules and their associated parameters because those rules are trained by the network.
4) It can extract most of the properties of the parking slot including location, orientation, type, and occupancy from a single AVM image.

## II. Related Research

Previous methods for detecting vacant parking spaces can be categorized into four approaches: slot marking-based, free space-based, user interface-based, and infrastructure-based [2]. The literature review of this paper focuses on the slot marking-based approach to which the proposed method belongs. Details of the other approaches can be found in our previous papers [2], [11], [14].

The slot marking-based approach detects parking spaces by recognizing markings on the ground. Its performance is not dependent on the existence and poses of the adjacent parked vehicles, but visually proper parking slot markings should be presented for this approach to work. All methods in this approach utilize vehicle-mounted cameras that can capture markings on the ground. The methods in [3],[4] detect parking slot markings in a semi-automatic manner. The method in [3] detects parking slot markings based on one manually designated point, and it was improved to detect various types of parking slot markings in [4] based on two manually designated points. The methods in [2],[5]-[18] detect parking slot markings in a full-automatic manner. The method in [5] detects parking slots based on color segmentation. The methods in [6]-[14] detect parking slot markings by finding lines using various techniques such as Hough transform, Radon transform, random sample consensus (RANSAC), or distance transform. The methods in [2], [15]-[18] detect parking slot markings by finding junctions of parking slots using a machine learning-based object detector or corner detector. Since this paper focuses on deep learning-based parking slot detection, those methods that use non-deep learning techniques are briefly introduced. A detailed introduction of them can be found in our previous papers [2], [11], [14].

Deep learning-based object detection methods have been widely researched because they show impressive detection performances for a variety of target objects under various conditions [19]. Deep learning-based object detection methods can be categorized into two approaches: two-stage and one-stage. The two-stage approach consists of two sequential steps. The first step generates category-independent region proposals and the second step recognizes classes of objects in the region proposals and refines their regions. Region-based CNN (RCNN) [23], Fast RCNN [24], Faster RCNN [25], RFCN (region-based fully convolutional network) [26], and mask RCNN [27] are representative methods in this approach. The two-stage approach has an advantage of high detection performance but has a limitation of slow detection speed. To mitigate this drawback, the one-stage approach has been suggested. This approach directly recognizes classes of objects along with their regions without generating region proposals. You only look once (YOLO) [28], single slot multibox detector (SSD) [29], and RetinaNet [30] are representative methods in this approach. The one-stage approach has an advantage of fast detection speed but has a limitation of relatively low detection performance compared to the two-stage approach.

As deep learning-based object detection has been actively researched, this technique has also been applied to the parking slot detection task [20]-[22]. The method in [20] generates entrances of parking slots by finding junctions using YOLO and pairing them using geometric rules. The generated entrances of parking slots are verified by a CNN and their orientations are estimated by a template matching technique. The method in [21] estimates locations and orientations of junctions using a customized CNN and pairs them using various geometric rules to detect parking slots. This method can only detect perpendicular and parallel parking slots, but not slanted parking slots. The method in [22] applies an existing two-stage object detection method (anchor free faster RCNN [31]) to the parking slot detection task. It finds four corners of the parking slot as a region proposal in the first step and refines the locations of the four corners while classifying the occupancy of the parking slot in the second step. The methods in [20] and [21] improved the parking slot detection performance using deep learning techniques along with various geometric rules. However, they cannot be trained end-to-end due to the use of manually designed geometric rules and require inconvenient process to manually set geometric rules and their associated parameters. The method in [22] can be trained end-to-end because it adapts the existing two-stage object detector. However, this method has limitations of detection performance and positioning accuracy because it simply adapts the method used for the general obstacle detection without specializing it for the parking slot detection task. Furthermore, this method incorrectly detects parking slots with certain orientations as mentioned in the paper, and its detection speed is relatively slow due to the used of the two-stage approach.

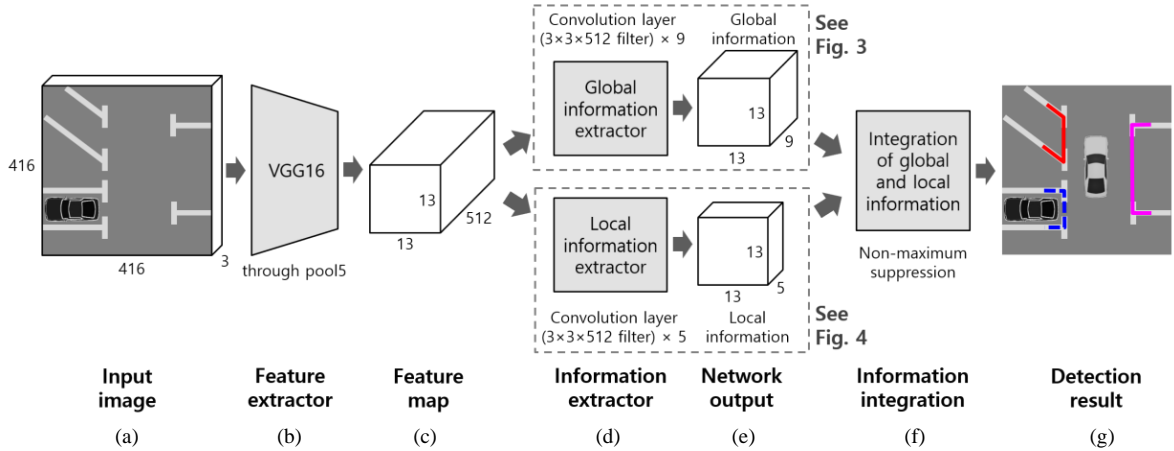

Fig. 2. Network architecture of the proposed method.

As a thorough literature review, it was found that this is the first paper that proposes an end-to-end trainable one-stage parking slot detection method. The proposed method has significant advantages over the previous deep learning-based parking slot detection methods as aforementioned at the end of the introduction.

## III. PROPOSED METHOD

### A. Network Architecture

This paper proposes an end-to-end trainable one stage parking slot detection method that simultaneously extracts the global and local information of the parking slot. To obtain such a detector, this paper suggests a novel network architecture as shown in Fig. 2. Fig. 2(a) indicates an input of the proposed network, which is a color AVM image with 416×416 pixels. Fig. 2(b) indicates a feature extractor. This paper uses VGG16 whose performance has been proven in various applications [32]. More sophisticated and recent networks have been tested but showed similar performances in terms of parking slot detection. The feature extractor can be changed depending on the application environment. VGG16 is used up to pool5, so that the dimension of the feature map obtained from it is 13×13×512 as shown in Fig. 2(c). Fig. 2(d) shows two extractors for the global and local information. The global and local information are represented by nine and five values, respectively. This paper extracts global and local information by applying only one convolution layer to the feature map obtained by VGG16 rather than applying multiple convolution layers to reduce computational costs. Thus, the global and local information extractors consist of nine and five 3×3×512 filters, respectively. Details on the global and local information extractors and their outputs will be described in the next section. The global and local information obtained by the proposed network are represented by a 13×13×9 tensor and a 13×13×5 tensor, respectively as shown in Fig. 2(e). The proposed method integrates the global and local information based on NMS as shown in Fig. 2(f) and produces a final parking slot detection result. Fig. 2(g) shows a conceptual example of the parking slot detection result. In this result, blue, magenta, and red lines indicate perpendicular, parallel, and slanted parking slots, respectively, and solid and dashed lines indicate vacant and occupied parking slots, respectively. The proposed method can extract most of the properties of the parking slot including location, orientation, type, and occupancy from a single AVM image.

### B. Information Extractors and Network Outputs

As shown in Fig. 2(e), the global and local information are represented by a 13×13×9 tensor and a 13×13×5 tensor, respectively. The spatial resolutions of those two tensors are 13×13. This means that the proposed method divides the input image into a grid of 13×13 cells and obtains global information and local information for each cell. Each cell is obliged to extract the global information of the parking slot including the cell center, and at the same time, it is obliged to extract the local information of the junction included in the cell. The reason for suggesting this architecture is that the cell inside a parking slot contains the overall information of the parking slot, and the cell including a junction contains the detailed positional information of the junction. Since each cell is used to obtain the global information of one parking slot and the location information of one junction, it is recommended that the cell size be set smaller than the minimum size of the parking slots and smaller than the minimum distance between two junctions. This allows one parking slot to include at least one cell and one cell to include at most one junction.

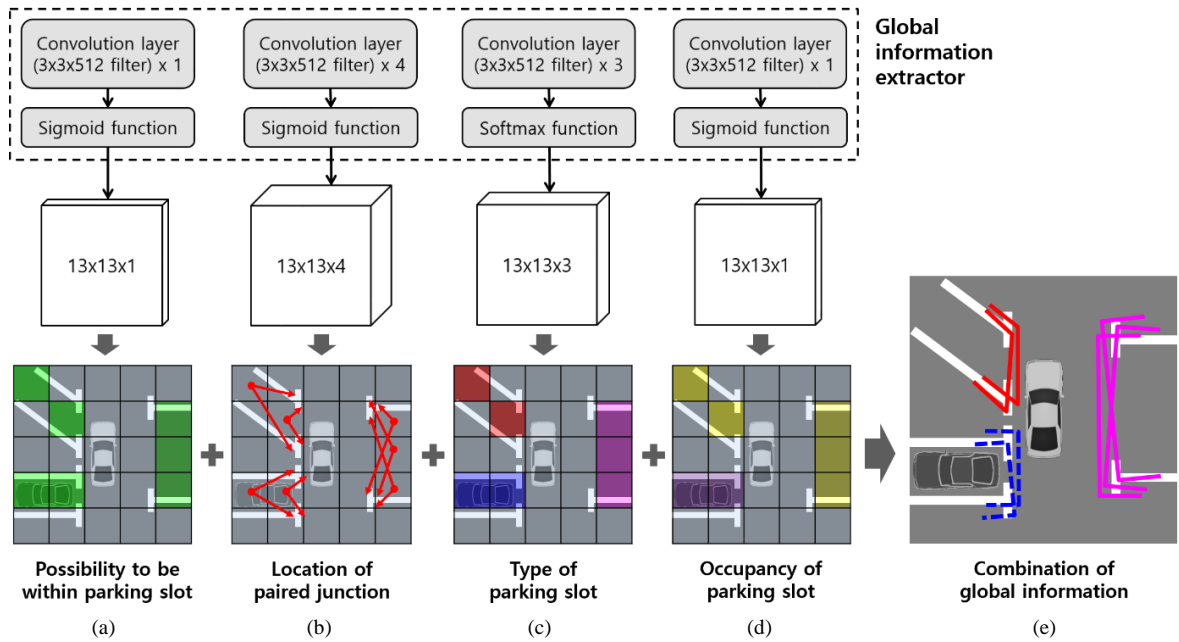

Fig. 3. Global information extractor and detailed global information obtained from it.

Fig. 3 and Fig. 4 show the detailed description of the information extractors and network outputs shown in Figs. 2(d) and (e). The global information extractor and its output (13×13×9 tensor) are divided into four parts (13×13×1, 13×13×4, 13×13×3, and 13×13×1 tensors) as shown in the first and second row of Fig. 3. In Fig. 3(a), the possibility that a cell center is included inside any parking slot is calculated. This is obtained by applying one 3×3×512 filter to the feature map shown in Fig. 2(c) followed by the sigmoid function, so that it is represented by a 13×13×1 tensor. The bottom of Fig. 3(a) shows its visual representation. In this figure, a cell whose center is included inside any parking slot has large value (green) and otherwise has small value (grey). the input image is divided into a grid of 5×5 cells for ease of understanding but divided into a grid of 13×13 cells in the actual implementation. In Fig. 3(b), the relative position from a cell center to paired junctions which represent the entrance of the parking slot including the cell center is calculated. This is obtained by applying four 3×3×512 filters to the feature map followed by the sigmoid function, so that it is represented by a 13×13×4 tensor. The bottom of Fig. 3(b) shows its visual representation. In this figure, only the results obtained from the cells located inside the parking slots are drawn. A pair of two red arrows indicates two 2D vectors connecting the cell center to the two junctions of the parking slot including the cell center. Since four values are needed to represent two 2D vectors for each cell, the dimension of the tensor is 13×13×4 as aforementioned. The position of each parking slot can be roughly estimated based on the two 2D vectors. In Fig. 3(c), the type of parking slot including a cell center is acquired. This is obtained by applying three 3×3×512 filters to the feature map followed by the softmax function, so that it is represented by a 13×13×3 tensor. In this paper, the parking slots are categorized into three types (perpendicular, parallel, and slanted) and those types are represented in one-hot encoding. Since three values are needed to represent three types for each cell, the dimension of the tensor is 13×13×3 as aforementioned. The bottom of Fig. 3(c) shows its visual representation. In this figure, blue, magenta, and red cells indicate perpendicular, parallel, and slanted parking slots, respectively. In Fig. 3(d), the occupancy of the parking slot including a cell center is acquired. This is obtained by applying one 3×3×512 filter to the feature map followed by the sigmoid function, so that it is represented by a 13×13×1 tensor. The bottom of Fig. 3(d) shows its visual representation. In this figure, a cell whose center is included inside the occupied parking slot has a large value (violet) and a cell whose center is included inside the vacant parking slot has a small value (yellow).

An intermediate parking slot detection result shown in Fig. 3(e) can be obtained by combining the whole global information shown in Figs. 3(a)-(d). In Fig. 3(e), blue, magenta, and red lines indicate perpendicular, parallel, and slanted parking slots, respectively, and solid and dashed lines indicate vacant and occupied parking slots, respectively. In this figure, two parking slot candidates are generated for the perpendicular and slanted parking slots because each of those slots includes two cells and three parking slot candidates are generated for the parallel parking slot because it includes three cells as shown in Figs. 3(a)-(d). The proposed method generates one parking slot candidate for each cell located inside the parking slot. It should be noted that the type and occupancy of the parking slot obtained by the global information extraction are accurate, but their positions are not. This is because the positions of the paired junction are estimated from the cell distant from those junctions. In the viewpoint of the automatic parking system, the positioning accuracy of the detection result is significantly important because the vehicle should be controlled based on the detected position. Therefore, this paper extracts the local information that includes precise location and orientation of the junction and uses them to improve the positioning accuracy of the parking slot.

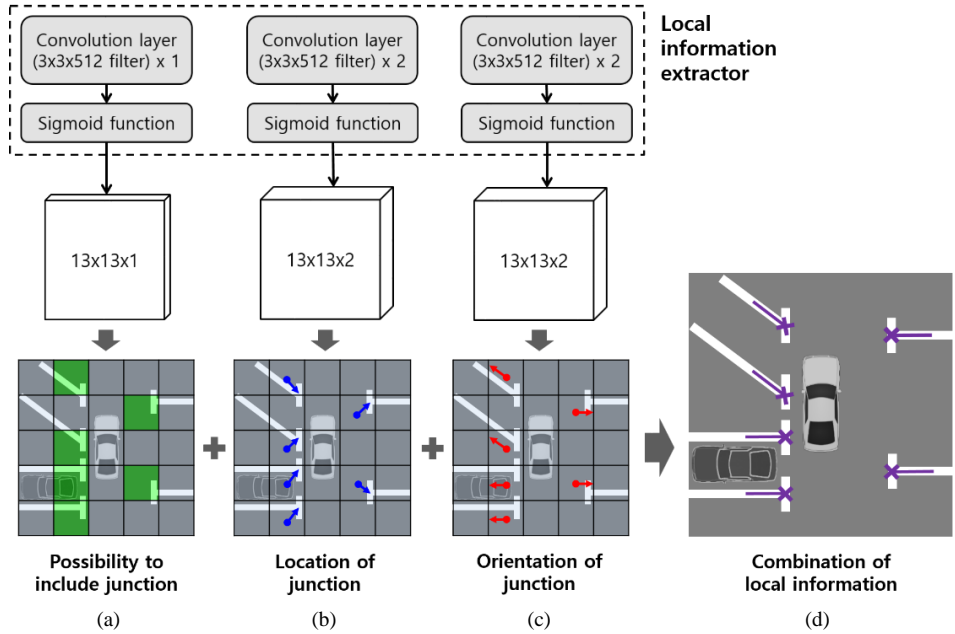

Fig. 4. Local information extractor and detailed local information obtained from it.

The local information extractor and its output ($13\times13\times5$ tensor) are divided into three parts ($13\times13\times1$, $13\times13\times2$, and $13\times13\times2$ tensor) as shown in the first and second row of Fig. 4. In Fig. 4(a), the possibility that a cell contains a junction is calculated. This is obtained by applying one $3\times3\times512$ filter to the feature map shown in Fig. 2(c) followed by the sigmoid function, so that it is represented by a $13\times13\times1$ tensor. The bottom of Fig. 4(a) shows its visual representation. In this figure, a cell containing a junction has large value (green), otherwise has small value (grey). In Fig. 4(b), the relative position from the cell center to the junction included in the cell is calculated. This is obtained by applying two $3\times3\times512$ filters to the feature map followed by the sigmoid function, so that it is represented by a $13\times13\times2$ tensor. The bottom of Fig. 4(b) shows its visual representation. In this figure, only the results obtained from the cells containing a junction are drawn. A blue arrow indicates a 2D vector connecting the cell center to the junction included in the cell. In Fig. 4(c), the orientation of the junction included in the cell is calculated. This is obtained by applying two $3\times3\times512$ filters to the feature map followed by the sigmoid function, so that it is represented by a $13\times13\times2$ tensor. The bottom of Fig. 4(c) shows its visual representation. A red arrow indicates a 2D vector that represents the orientation of the junction included in the cell. Only the direction of this vector is estimated and used.

A junction detection result shown in Fig. 4(d) can be obtained by combining the whole local information shown in Figs. 4(a)-(c). In Fig. 4(d), violet crosses and lines indicate the locations and orientations of the detected junctions, respectively. It should be noted that the positions of the junctions obtained in the local information extraction are more accurate than those obtained in the global information extraction. This is because the position of the junction is estimated from the cell including the junction during the local information extraction, but it is estimated from the cell distant from the junction during the global information extraction. Therefore, the proposed method uses the precise junction position acquired by the local information extraction to improve the positioning accuracy of the parking slot acquired by the global information extraction.

### C. Integration of Global and Local Information

Final parking slot detection results are produced by integrating the global and local information. The information integration is simply performed by a junction-based NMS. In this step, if the junction obtained in the global information extraction (global junction) exists near the junction obtained in the local information extraction (local junction), the global junction is replaced by the local junction because the position of the local junction is more precise than that of the global junction. Through the junction-based NMS, the parking slots of the global information are matched with the junctions of the local information, so that the global and local information are integrated together. After the integration, the orientation of the parking slot is set to 90° with respect to the line connecting two junctions in case of the perpendicular or parallel parking slot and is set the average of the orientations of two junctions in case of the slanted parking slot. Figs. 5(a) and (b) show the global and local information, respectively, and Fig. 5(c) shows the junction-based NMS result. In Fig. 5(c), black crosses indicate the global junctions replaced by the local junctions. Two global junctions of one slanted parking slot (lower red line in Fig. 5(a)) are replaced by two local junctions located nearby as shown in Fig. 5(c), and two global junctions of the other slanted parking slot (upper red line in Fig. 5(a)) are not replaced by the local junctions as shown in Fig. 5(c) because local junctions are not located near them. All global junctions of two perpendicular parking slots (blue lines in Fig. 5(a)) are replaced by local junctions, so that their positions become identical as shown in Fig. 5(c). In this figure, ×2 indicates that two parking slots overlap each other. Global junctions of two parallel parking slots (two upper magenta lines in Fig. 5(a)) are replaced by local junctions, so that their positions become identical as shown in Fig. 5(c). But in the case of the other parallel parking slot (lower magenta line in Fig. 5(a)), only one of its global junctions is replaced by a nearby local junction as shown in Fig. 5(c). After the junction-based NMS, a slot-based NMS is performed. This step first eliminates the parking slots in which all their global junctions are not replaced by the local junctions. In Fig. 5(c), one slanted parking slot (the upper red line) and one parallel parking slot (lower magenta line) are eliminated. If there are overlapping parking slots, only the one that has the highest possibility calculated in Fig. 3(a) is selected. Fig. 5(d) shows the final parking slot detection result after the slot-based NMS.

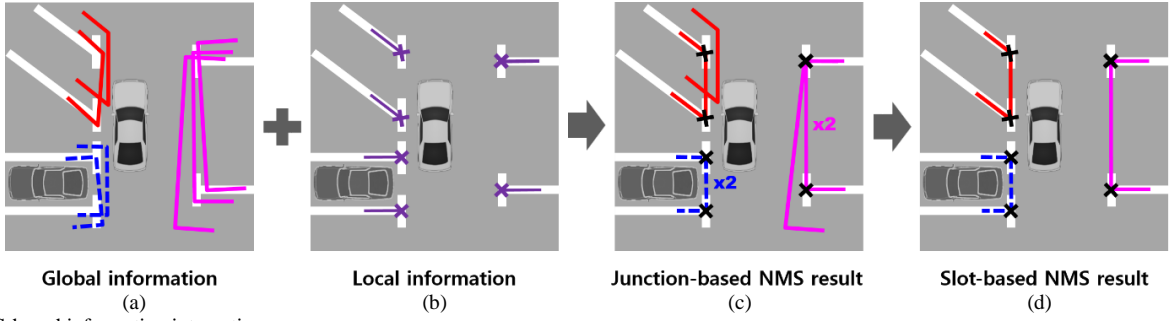

Fig. 5. NMS-based information integration.

### D. Training

The feature extractor was initialized by the weights pre-trained on ImageNet and 14 convolution filters used for extracting the global and local information were initialized by Xavier uniform initializer. All weights were optimized by Adam optimizer whose learning rate, $\beta_1$, $\beta_2$, and $\varepsilon$ were set to $10^{-4}$, 0.9, 0.999, and $10^{-8}$, respectively. The proposed network was trained for 100 epochs and the batch size was set to 24. Fig. 6 shows the training procedure of the proposed network. As shown in Fig. 6(g), the ground truth of the parking slot is divided into seven parts. Four parts are used as the ground truth of the global information and three parts are used as the ground truth of the local information. Those ground truths are compared to the network outputs (global and local information in Fig. 6(e)) to calculate losses as shown in Fig. 6(f).

The final loss is calculated by the weighted sum of the seven losses that correspond to seven ground truths in Fig. 6(g). Four losses ($loss_{sp}$, $loss_{sxy}$, $loss_{st}$, and $loss_{so}$) are for the global information and three losses ($loss_{jp}$, $loss_{jxy}$, and $loss_{jv}$) are for the local information as

$$loss = \underbrace{w_{sp} \cdot loss_{sp} + w_{sxy} \cdot loss_{sxy} + w_{st} \cdot loss_{st} + w_{so} \cdot loss_{so}}_{\text{Losses for global information}} \\ + \underbrace{w_{jp} \cdot loss_{jp} + w_{jxy} \cdot loss_{jxy} + w_{jv} \cdot loss_{jv}}_{\text{Losses for local information}} \quad (1)$$

where $w_{sp}$, $w_{sxy}$, $w_{st}$, $w_{so}$, $w_{jp}$, $w_{jxy}$, and $w_{jv}$ are the weights for the seven losses, and they are experimentally set to 40, 170, 0.05, 3, 300, 3000, and 1000, respectively, based on their magnitudes in the training dataset. Each loss will be explained in detail one by one. It should be noted that all losses are designed by considering that all network outputs are between 0 and 1.

The loss for the possibility that a cell center is included in any parking slot, $loss_{sp}$ is calculated as

$$loss_{sp} = \sum_{i=1}^{S^2} \left[ I^i_{slot} \left( sp^i_{pred} - sp^i_{true} \right)^2 + \lambda_{slot} \left( 1 - I^i_{slot} \right) \left( sp^i_{pred} - sp^i_{true} \right)^2 \right] \quad (2)$$

where $sp^i_{true}$ is the ground truth for the possibility that the center of the $i$-th cell is included in any parking slot. It is set to 1 if included or 0 if not. The input image is assumed to be divided into a grid of $S \times S$ cells. $sp^i_{pred}$ is the prediction result of the network for $sp^i_{true}$. $I^i_{slot}$ indicates whether the center of the $i$-th cell is included in any parking slot and is set to 1 if included or 0 if not. $\lambda_{slot}$ is added to compensate for the unbalance between the number of cells included in the parking slots and the number of cells that are not. This is set to 0.2 based on the ratio of those numbers in the training dataset.

The loss for the relative position from the center to the paired junctions, $loss_{sxy}$ is calculated as

$$loss_{sxy} = \sum_{i=1}^{S^2} I^i_{slot} \left[ \left\{ 2\left( sx1^i_{pred} - 0.5 \right) - sx1^i_{true}/L_{max} \right\}^2 \\ + \left\{ 2\left( sy1^i_{pred} - 0.5 \right) - sy1^i_{true}/L_{max} \right\}^2 \\ + \left\{ 2\left( sx2^i_{pred} - 0.5 \right) - sx2^i_{true}/L_{max} \right\}^2 \\ + \left\{ 2\left( sy2^i_{pred} - 0.5 \right) - sy2^i_{true}/L_{max} \right\}^2 \right] \quad (3)$$

where $(sx1^i_{true}, sy1^i_{true})$ and $(sx2^i_{true}, sy2^i_{true})$ are two 2D vectors that represent the ground truth for the relative position from the center of the $i$-th cell to the paired junctions of the parking slot including the $i$-th cell. These values are divided by $L_{max}$ and normalized to the values between -1 and 1. $L_{max}$ is set to the maximum length of the parking slot marking, which is $7m \times$PPM in this paper. PPM stands for pixel per meter and is calculated from the AVM image calibration procedure. ($sx1^i_{pred}$, $sy1^i_{pred}$) and ($sx2^i_{pred}$, $sy2^i_{pred}$) are the prediction results of the network for ($sx1^i_{true}$, $sy1^i_{true}$) and ($sx2^i_{true}$, $sy2^i_{true}$), respectively. Since they are between 0 and 1, we subtract 0.5 from them and multiply by 2 to make them between -1 and +1, which is the range of the corresponding ground truth.

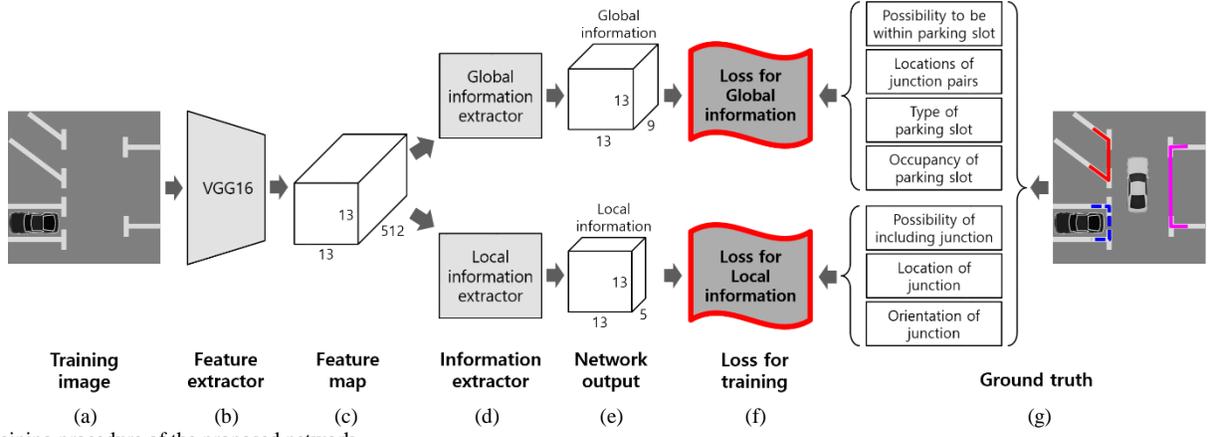

Fig. 6. Training procedure of the proposed network.

The loss for the parking slot type, $loss_{st}$ is calculated based on the categorical cross-entropy as

$$loss_{st} = \sum_{i=1}^{S^2} I^i_{slot} \left[ -\sum_{c=1}^{3} \left\{ \lambda_{st,c} st^i_{true,c} \log\left(st^i_{pred,c}\right) \right\} \right] \quad (4)$$

where $st^i_{true,c}$ is the ground truth for the probability that the type of parking slot containing the center of the $i$-the cell is $c$ (1, 2, or 3). This is represented in one-hot encoding. That is, if the type of the parking slot is perpendicular, parallel or slanted, ($st^i_{true,1}$, $st^i_{true,2}$, $st^i_{true,3}$) is set to (1,0,0), (0,1,0), or (0,0,1), respectively. $st^i_{pred,c}$ is the prediction result of the network for $st^i_{true,c}$. $\lambda_{st,c}$ is added to compensate for the unbalance among the numbers of the cells included in different types of the parking slots. $\lambda_{st,1}$, $\lambda_{st,2}$, and $\lambda_{st,3}$ are set to 1.76, 2.86, and 31.65, respectively, based on the ratio of those numbers in the training dataset.

The loss for the parking slot occupancy of the cell, $loss_{so}$ is calculated as

$$loss_{so} = \sum_{i=1}^{S^2} \left[ \lambda_{occ} I^i_{occ} \left(so^i_{pred} - so^i_{true}\right)^2 + \lambda_{vac} I^i_{vac} \left(so^i_{pred} - so^i_{true}\right)^2 \right] \quad (5)$$

where $so^i_{true}$ is the ground truth for the occupancy of the parking slot that includes the $i$-th cell. It is set to 1 if occupied or 0 if vacant. $so^i_{pred}$ is the prediction result of the network for $so^i_{true}$. $I^i_{occ}$ indicates whether the center of the $i$-th cell is included in the occupied parking slot and is set to 1 if included or 0 if not. $I^i_{vac}$ indicates whether the center of the $i$-th cell is included in the vacant parking slot and is set to 1 if included or 0 if not. $\lambda_{occ}$ and $\lambda_{vac}$ are added to compensate for the unbalance between the number of cells included in the occupied parking slots and the number of cells included in the vacant parking slots. They are set to 1.55 and 3.30, respectively, based on the ratio of those numbers in the training dataset.

The loss for the possibility that the cell contains a junction, $loss_{jp}$ is calculated as

$$loss_{jp} = \sum_{i=1}^{S^2} \left[ I^i_{junc} \left( jp^i_{pred} - jp^i_{true} \right)^2 + \lambda_{junc} \left(1 - I^i_{junc}\right) \left( jp^i_{pred} - jp^i_{true} \right)^2 \right] \quad (6)$$

where $jp^i_{true}$ is the ground truth for the possibility that the $i$-th cell is a junction. It is set to 1 if contains or 0 if not. $jp^i_{pred}$ is the prediction result of the network for $jp^i_{true}$. $I^i_{junc}$ indicates whether the $i$-th cell contains a junction and is set to 1 if contains or 0 if not. $\lambda_{junc}$ is added to compensate for the unbalance between the number of cells that contains junctions and the number of cells that do not. It is set to 0.02 based on the ratio of those numbers in the training dataset.

The loss for the relative position from the cell center to the junction included in the cell, $loss_{jxy}$ is calculated as

$$loss_{jxy} = \sum_{i=1}^{S^2} I^i_{junc} \left[ \left\{ \left(jx^i_{pred} - 0.5\right) - jx^i_{true}/W_{cell} \right\}^2 + \left\{ \left(jy^i_{pred} - 0.5\right) - jy^i_{true}/H_{cell} \right\}^2 \right] \quad (7)$$

where ($jx^i_{true}$, $jy^i_{true}$) is a 2D vector that represents the ground truth for the relative position from the center of the $i$-th cell to the junction included in the $i$-th cell. These values are divided by $W_{cell}$ and $H_{cell}$ and normalized to the values between -0.5 and +0.5. Since the input (416×416 pixels) is divided into 13×13 cells, both $W_{cell}$ and $H_{cell}$ are 32. ($jx^i_{pred}$, $jy^i_{pred}$) is the prediction result of the network for ($jx^i_{true}$, $jy^i_{true}$). Since they are between 0 and 1, we subtract 0.5 from them to make them between -0.5 and +0.5, which is the range of the corresponding ground truth.

The loss for the orientation of the junction included in the cell, $loss_{jv}$ is calculated as

$$loss_{jv} = \sum_{i=1}^{S^2} I^i_{junc} \left[ \left\{ 2\left(jvx^i_{pred} - 0.5\right) - jvx^i_{true} \right\}^2 + \left\{ 2\left(jvy^i_{pred} - 0.5\right) - jvy^i_{true} \right\}^2 \right] \quad (8)$$

where ($jvx^i_{true}$, $jvy^i_{true}$) is a 2D normal vector that represents the ground truth for the orientation of the junction included in the $i$-th cell. ($jvx^i_{pred}$, $jvy^i_{pred}$) is the prediction result of the network for ($jvx^i_{true}$, $jvy^i_{true}$). Since they are between 0 and 1, we subtract 0.5 from them and multiply by 2 to make them between -1 and +1, which is the range of the corresponding ground truth.

TABLE I
SUMMARY OF THE DATASET (PS2.0)

| Data | | Training | Test |
|---|---|---|---|
| Images | | 9827 | 2338 |
| Slots | Perpendicular | 5668 | 936 |
| | Parallel | 3492 | 1151 |
| | Slanted | 316 | 81 |
| | Total | 9476 | 2168 |

TABLE II
PARKING SLOT DETECTION PERFORMANCE EVALUATION AND COMPARISON

| Method | #False Negative | #False Positive | Recall | Precision |
|---|---|---|---|---|
| **Proposed method** | **5** | **5** | **99.77%** | **99.77%** |
| DeepPS [20] | 22 | 8 | 98.99% | 99.63% |

TABLE III
POSITION ACCURACY EVALUATION AND COMPARISON

| Method | Location error (pixel) | | Orientation error (degree) | |
|---|---|---|---|---|
| | mean | std. | mean | std. |
| **Proposed method** | **1.02** | **0.72** | **0.18** | **0.30** |
| DeepPS [20] | 1.09 | 0.74 | 0.39 | 0.57 |

TABLE IV
TYPE CLASSIFICATION PERFORMANCE OF THE PROPOSED METHOD

| Type | #Correctly detected slots | #Correctly classified slots | Classification rate (%) |
|---|---|---|---|
| Perpendicular | 934 | 934 | 100% |
| Parallel | 1151 | 1151 | 100% |
| Slanted | 78 | 78 | 100% |
| Total | 2163 | 2163 | 100% |

TABLE V
OCCUPANCY CLASSIFICATION PERFORMANCE OF THE PROPOSED METHOD

| Occupancy | #Correctly detected slots | #Correctly classified slots | Classification rate (%) |
|---|---|---|---|
| Vacant | 1609 | 1597 | 99.25% |
| Occupied | 554 | 551 | 99.46% |
| Total | 2163 | 2148 | 99.31% |

## IV. EXPERIMENTS

### A. Dataset

The proposed method was quantitatively evaluated using the publicly available AVM image dataset called Tongji Parking Slot Dataset 2.0 (PS2.0) [33]. This dataset consists of 9827 training images with 9476 parking slots and 2338 test images with 2168 parking slots and includes three types of parking slot markings (perpendicular, parallel, and slanted). Table 1 shows details of the dataset. This dataset contains images taken under various illumination conditions including outdoors and indoors, daytime and nighttime, sunny and rainy days, strong shadows, etc. The original AVM image includes 10×10$m$ around the vehicle and its resolution is 600×600 pixels. The proposed method resizes the original image to 416×416 pixels and feeds it into the network as an input image. This dataset contains the ground truth positions of the parking slots and junctions. Because it does not contain the ground truth of the occupancies of the parking slots, we manually designated them. While adding this ground truth, a parking slot that includes the ego-vehicle region is labeled as vacant because the ego-vehicle can park in that slot.

### B. Performance Evaluation and Comparison

The proposed method was quantitatively evaluated using the evaluation criteria provided by PS2.0 [33]. According to the criteria, a parking slot is considered as a true positive if its two junctions are within 12 pixels from their ground truth locations and its orientation is within 10° from the ground truth orientation. All detected parking slots that do not meet these conditions are considered as false positives. For the performance evaluation and comparison, recall and precision are calculated as

$$\text{recall} = \frac{\text{\# True Postive}}{\text{\# True Postive + \# False Negative}}$$
$$\text{precision} = \frac{\text{\# True Postive}}{\text{\# True Postive + \# False Postive}} \quad (9)$$

Table 2 shows the parking slot detection performances of the proposed method and DeepPS [20]. The detection result of DeepPS is obtained from the publicly available code released by its authors [33]. Since performances of the previous methods based on non-deep learning techniques are much inferior to the proposed method and DeepPS, they are not presented in Table 2. Their performances can be found in [20]. In addition, the other two deep learning-based methods suggested in [21] and [22] are not compared with the proposed method and DeepPS because they have critical drawbacks that they cannot detect slanted parking slots or parking slots with certain orientations, respectively, as mentioned in their papers. DeepPS is the only previous method based on the deep learning technique that can handle all situations included in PS2.0. As shown in Table 2, the proposed method misses only five parking slots out of 2168 and produces five false positives while DeepPS misses 22 parking slots and produces eight false positives. The recall and precision of the proposed method are all 99.77%. The reason that the proposed method outperforms DeepPS is as follows: The whole process of DeepPS is difficult to be integratedly optimized because it is a combination of the CNN trained by the data and the rules designed by hands. Contrarily, the proposed method can be integratedly optimized because it is designed as an end-to-end trainable fully convolutional network. In addition, DeepPS obtains the location of the parking slot using the CNN-based object detector (YOLO) but obtains the orientation of the parking slot using the template matching technique. Contrarily, the proposed method obtains both the location and orientation of the parking slot using the CNN, so that it can provide more accurate parking slots compared to DeepPS. Table 3 shows the positioning accuracies of the proposed method and DeepPS. These accuracies are obtained from only the correctly detected parking slots. It can be noticed that the location errors of the two methods are similar because both methods use the CNN for the junction detection, but the orientation error of the proposed method is less than that of DeepPS because the proposed method estimates the orientation of the parking slot using the CNN while DeepPS uses the template matching.

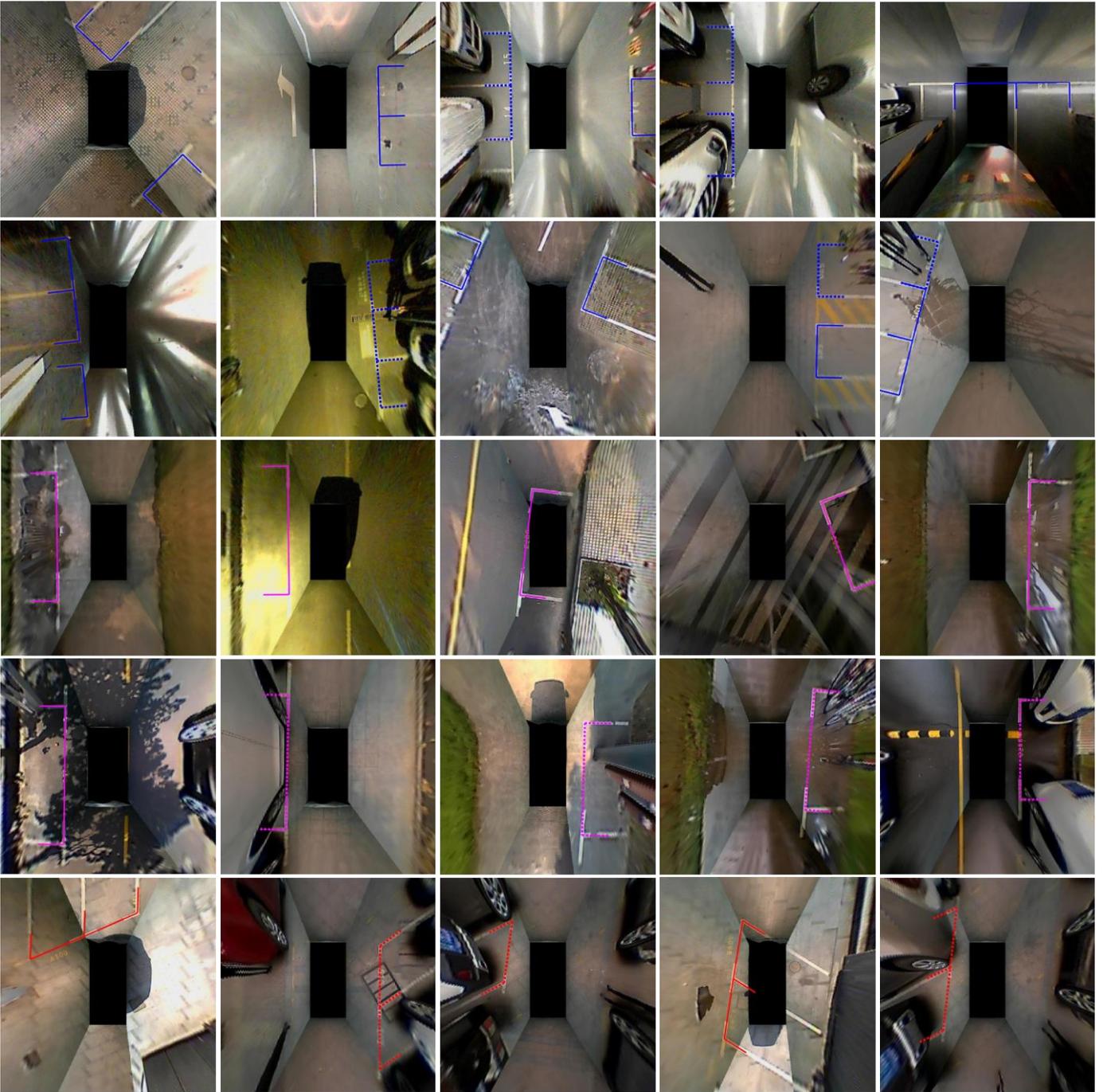

Fig. 7. Parking slot detection results of the proposed method under various conditions. Blue, magenta, and red lines indicate perpendicular, parallel, and slanted parking slots, respectively, and solid and dotted lines indicate vacant and occupied parking slots, respectively.

Unlike DeepPS, the proposed method not only estimates the positions of the parking slots but also classifies their types and occupancies. Table 4 shows the type classification performance of the proposed method. It correctly classifies the types of all 2163 correctly detected parking slots, so that its type classification accuracy is 100%. The parking slot type can be obtained using the handcrafted geometric features such as width, height, and orientation of the detected parking slot. However, this paper shows that it can be easily obtained by the

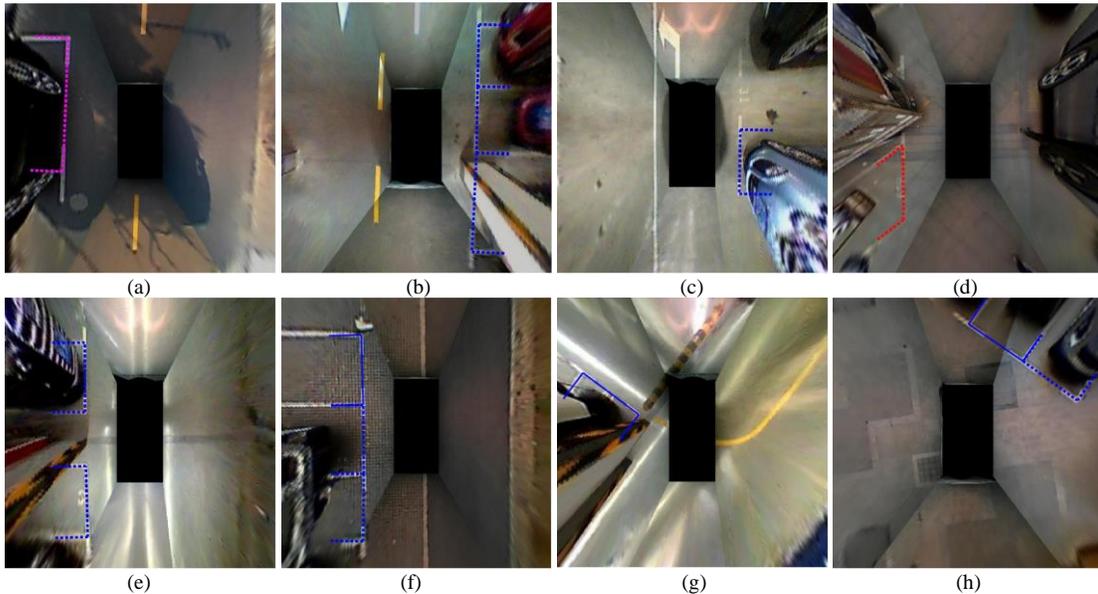
Fig. 8. Failure cases of the proposed method. (a) and (b) show false positives, (c) and (d) show false negatives, and (e)-(h) show occupancy classification errors.

CNN along with the detection result without using any handcrafted features. Table 5 shows the occupancy classification performance of the proposed method. It correctly classifies occupancies of 2148 parking slots out of 2163, so that its occupancy classification accuracy is 99.31%. In Tables 2-5, it is shown that the proposed method can extract most of the properties of the parking slot including location, orientation, type, and occupancy from a single AVM image using an end-to-end trainable fully convolutional network.

Fig. 7 shows the parking slot detection results of the proposed method in various situations included in the test dataset of PS2.0. In this figure, blue, magenta, and red lines indicate perpendicular, parallel, and slanted parking slots, respectively, and solid and dotted lines indicate vacant and occupied parking slots, respectively. It can be noticed that the proposed method not only detects the parking slots but also recognizes their types and occupancies under a variety of road conditions (reflected lights, standing water, strong shadows, stains, asphalt, concrete, bricks, etc.), illumination conditions (outdoors and indoors, daytime and nighttime, sunny and rainy days, etc.), and obstacle conditions (cars, pillars, bicycles, pedestrians, etc.).

Fig. 8 shows failure cases of the proposed method. Figs. 8(a) and (b) show false detections. In Fig. 8(a), this method detects the parking slot whose one junction is occluded by the parked vehicle. This detection may be a true positive, but it was considered as a false positive because the ground truth of PS2.0 dataset includes only the parking slots whose junctions are all visible. This failure case appears three times. In Fig. 8(b), this method produces a false positive by detecting a space where a pillar exists between two parking slots (lower blue line). Figs. 8(c) and (d) show miss detections. This method cannot detect two parking slots in Figs. 8(c) and (d) due to the severely faded parking slot marking and heavy occlusion of one junction, respectively. Figs. 8(e)-(h) show occupancy classification failures. In Figs. 8(e) and (f), two parking slots at the bottom are incorrectly classified as occupied because of the stretched image regions of the adjacent pillar and parked car, respectively. In Figs. 8(g) and (h), two parking slots are incorrectly classified as vacant.

The proposed method was implemented using Python with Keras. NVIDIA GEFORCE GTX 1080Ti was used for the experiment. The inference phase of the proposed method requires 16.66*ms* to process one image, so that it can process 60 images per second. DeepPS was implemented using C++ with Darkent and Caffe. It requires 23.83*ms* to process one image using the same GPU. It is difficult to directly compare those two methods because they use different frameworks, but in general, the proposed method is expected to be faster than DeepPS considering that Python with Keras is slower than C++ with Darknet and Caffe.

## V. CONCLUSIONS

This paper proposes an end-to-end trainable one-stage parking slot detection method. The proposed method has several obvious advantages over the previous deep learning-based parking slot detection methods. First, it can achieve both high detection rate and positioning accuracy by integrating the global and local information of the parking slot. Second, it can be trained end-to-end and rapidly detect parking slots because it uses a fully convolutional network and one-stage detection strategy. Third, it does not require inconvenient procedures for setting geometric rules and their associated parameters because those rules are trained by the network. Last, it can extract most of the properties of the parking slot including location, orientation, type, and occupancy from a single AVM image. Experimental results showed that the proposed method outperforms previous methods while requiring a small amount of computational cost. In the future, we are planning to compress the proposed network and embed it into recently released edge artificial intelligence (AI) chips.